\documentclass[9pt,shortpaper,twoside,web]{ieeecolor}
\usepackage{generic}
\usepackage{cite}
\usepackage{amsmath,amssymb,amsfonts}
\usepackage{algorithmic}
\usepackage{graphicx}
\usepackage{textcomp}

\usepackage{booktabs}
\usepackage{multirow}
\usepackage[normalem]{ulem} 
\newcommand{\ve}[1]{\mathbf{#1}}
\newcommand{\vv}[1]{\mbox{\boldmath $#1$}}
\newcommand{\tabincell}[2]{\begin{tabular}{@{}#1@{}}#2\end{tabular}} 
\usepackage{lipsum}
\usepackage{subfigure}
\usepackage{bbm}
\usepackage{algorithm}
\usepackage{algorithmic}

\def\BibTeX{{\rm B\kern-.05em{\sc i\kern-.025em b}\kern-.08em
    T\kern-.1667em\lower.7ex\hbox{E}\kern-.125emX}}
\begin{document}
\title{PCCT: Progressive  Class-Center Triplet Loss for Imbalanced Medical Image Classification}
\author{Kanghao Chen, Weixian Lei, Rong Zhang, Shen Zhao, Wei-shi Zheng, Ruixuan Wang
\thanks{This work is supported in part by the National Natural Science Foundation of China (grant No. 62071502, U1811461).
}
\thanks{Kanghao Chen, Weixian Lei, Rong Zhang, Wei-shi Zheng and Ruixuan Wang are with are School of Computer Science and Engineering, and Shen Zhao is with the School of Intelligent Systems Engineering, Sun Yat-Sen University, Guangzhou, China.}
\thanks{Kanghao Chen and Weixian Lei contribute equally to this work.}
\thanks{Corresponding authors: Ruixuan Wang (wangruix5@ mail.sysu.edu.cn), Shen Zhao (zhaosh35@mail.sysu.edu.cn).}
}


\maketitle

\begin{abstract}
Imbalanced training data is a significant challenge for medical image classification. In this study, we propose a novel Progressive Class-Center Triplet (PCCT) framework to alleviate the class imbalance issue particularly for  diagnosis of rare diseases, mainly by carefully designing the triplet sampling strategy and the triplet loss formation. Specifically, the PCCT framework includes two successive stages. In the first stage, PCCT trains the diagnosis system via a class-balanced triplet loss to coarsely separate distributions of different classes. In the second stage, the PCCT framework further improves the diagnosis system via a class-center involved triplet loss to cause a more compact distribution for each class. For the class-balanced triplet loss, triplets are sampled equally for each class at each training iteration, thus alleviating the imbalanced data issue. For the class-center involved triplet loss, the positive and negative samples in each triplet are replaced by their corresponding class centers, which enforces data representations of the same class closer to the class center. Furthermore, the class-center involved triplet loss is extended to the pair-wise ranking loss and the quadruplet loss, which demonstrates the generalization of the proposed framework. Extensive experiments support that the PCCT framework works effectively for medical image classification with imbalanced training images. On two skin image datasets and one chest X-ray dataset, the proposed approach respectively obtains the mean F1 score 86.2, 65.2, and 90.66 over all classes and 81.4, 63.87, and 81.92 for rare classes, achieving state-of-the-art performance and outperforming the widely used methods for the class imbalance issue. 
\end{abstract}

\begin{IEEEkeywords}
Data imbalance, Medical image classification, Triplet loss.
\end{IEEEkeywords}

\section{Introduction}
\label{sec:introduction}
\IEEEPARstart{C}{lass} imbalance issue is ubiquitous in medical diagnosis~\cite{tschandl2018ham10000} because large-scale clinical datasets often exhibit imbalanced class distributions.
For example, in clinical diagnosis, the data is by nature heavily imbalanced~\cite{Shilaskar2019DiagnosisSF} because common diseases occur more frequently than rare disease.   This raises a major challenge for modern deep learning models because most of them assume balanced class distributions in the training dataset.
When presented with an imbalanced dataset, the training procedure is dominated by frequent classes, and the trained model tends to perform better on these frequent classes but significantly worse on infrequent classes~\cite{Jain2017AddressingCI}.


Many approaches attempt to solve the class imbalance issue. 
For example, the re-sampling~\cite{Buda2018Systematic} strategy can be applied to  over-sample the limited data from infrequent classes or under-sample the data from frequent classes to balance training data across classes.
The re-weighting~\cite{Cui2019CBloss} strategy sets larger weights to the loss terms related to infrequent classes, which makes balanced loss terms across classes. 
Still relevant to modification of training loss, the traditional margin-based loss may be refined by setting smaller margin for frequent classes and larger margin for infrequent classes to alleviate the class-imbalance issue~\cite{Cao2019LDAM}.
Another approach~\cite{Lin2017FocalLoss} adaptively sets a higher weight for the sample that is difficult to recognize.
Besides the re-balancing strategies mentioned above,
other strategies have also been proposed by improving the representation ability of the deep neural network.
These can be achieved by class-imbalanced representation learning, such as transfer learning~\cite{Liu2019LargeScaleLR}, semi-supervised and self-supervised learning~\cite{Yang2020RethinkingTV}.
Furthermore, the representation learning can be combined with existing re-balancing strategies, e.g., by first performing representation learning of the feature extractor and then applying re-balancing strategy to the model output side~\cite{Kang2020DecouplingRA} or the input side~\cite{Zhou2020BBNBN}.
Although the above-mentioned approaches can alleviate the class imbalance issue to some extent, these strategies still cause difficulties and adverse effects in practice.
For example,
re-sampling and re-weighting have the risk of over-inverting the data frequency of infrequent classes, which would unexpectedly damage the overall representation ability of the learned features~\cite{Zhou2020BBNBN}.
%

Different from the above model training approaches, triplet loss~\cite{schroff2015facenet, hermans2017defense,song2018mask, he2018triplet, Ge2018DeepML} is another popular scheme for representation learning.
Triplet loss~\cite{schroff2015facenet} is originally proposed to capture face embedding for face recognition and verification, and then triplet loss and its variants~\cite{hermans2017defense,he2018triplet} are adopted to train the feature extractor for more applications such as person re-identification and 3D shape retrieval. 
Furthermore, region-level triplet loss~\cite{song2018mask} is proposed to capture discriminative and robust features,
and the adaptively updated hierarchical tree~\cite{Ge2018DeepML} is applied to sample more informative triplets to train the feature extractor.
However, current triplet losses have not been attempted for solving the class imbalance issue, because the triplets whose anchors come from frequent classes will be much more than the triplets whose anchors come from infrequent classes. 
In addition, triplet losses may suffer from instability during training with large-scale datasets where large visual differences may often appear in samples from the same class.


In this study, by extending the triplet loss, we propose a novel two-stage Progressive Class-Center Triplet (PCCT) framework to alleviate the class imbalance issue for diagnosing rare diseases. 
The basic idea is to first use class-balanced triplets to train the feature extractor, and then replace the positive and negative samples in each triplet with their corresponding class centers to further fine-tune the feature extractor. 
With the help of the two-stage training process, the proposed framework captures balanced distributions for different classes in the feature space.
The main contributions of this study are summarized as follows.
\begin{itemize}
    \item We propose a novel two-stage Progressive Class-Center Triplet (PCCT) framework  to capture discriminative and class-compact representation, which is effective to solve the class imbalance issue in clinical diagnosis.
    \item We design a class-balanced triplet sampler in the first stage to alleviate the class imbalance issue and propose a class-center involved triplet loss in the second stage to encourage compact representation for each class.
    \item We evaluate our approach on three imbalanced medical image datasets for disease diagnosis. The proposed approach consistently achieves state-of-the-art performance and outperforms the widely used methods for the class-imbalance issue.
\end{itemize}
Note that this work is an extension of the previous conference publication~\cite{lei2020class} in the following aspects.
\begin{enumerate}
    \item  We delve into the two-stage training process of our PCCT. More detailed analyses of the relationship between the two stages and extensive experiments are performed, clearly demonstrating the effectiveness of the novel training process in the imbalanced medical diagnosis.
    \item  We additionally propose to directly train class centers together with the feature extractor, allowing much less computation cost during training but equivalent classification performance, which is referred to as Efficient-PCCT.
    In comparison, the class center of each class is originally estimated by the mean feature representations over all training samples from the same class, which is time-consuming. 
    \item We extend the class-center involved loss to other metric learning, including the pair-wise ranking loss and the quadruplet loss. Corresponding empirical evaluations  show that the class-center involved metric learning outperforms the original learning strategies.
    \item More extensive evaluations have been included, not only on the two skin image datasets but also on the new chest X-ray dataset.
\end{enumerate}

%


\section{Methods}
\begin{figure*}[t]
\vspace{-5mm}
	\setlength{\abovecaptionskip}{2 mm}
    \begin{center}
       \subfigure[Overview]{
		
                        \includegraphics[width=\linewidth]{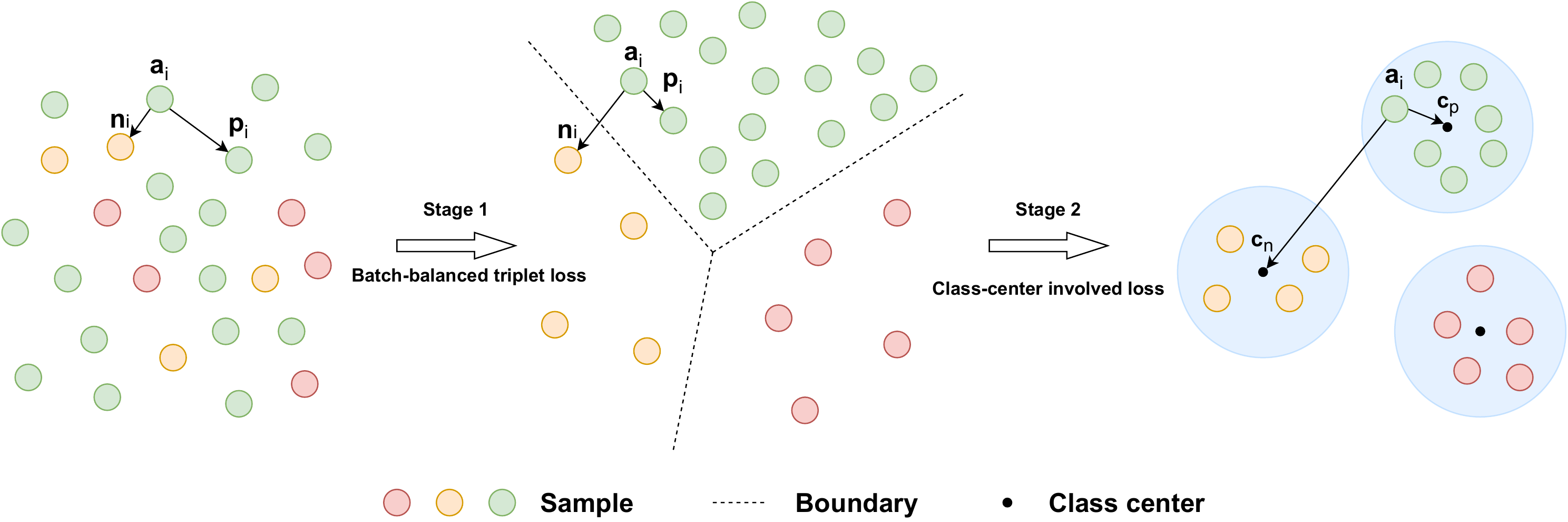}
	}
	\\
			

 \subfigure[Batch-balanced triplet sampler]{\includegraphics[width=4.5in]{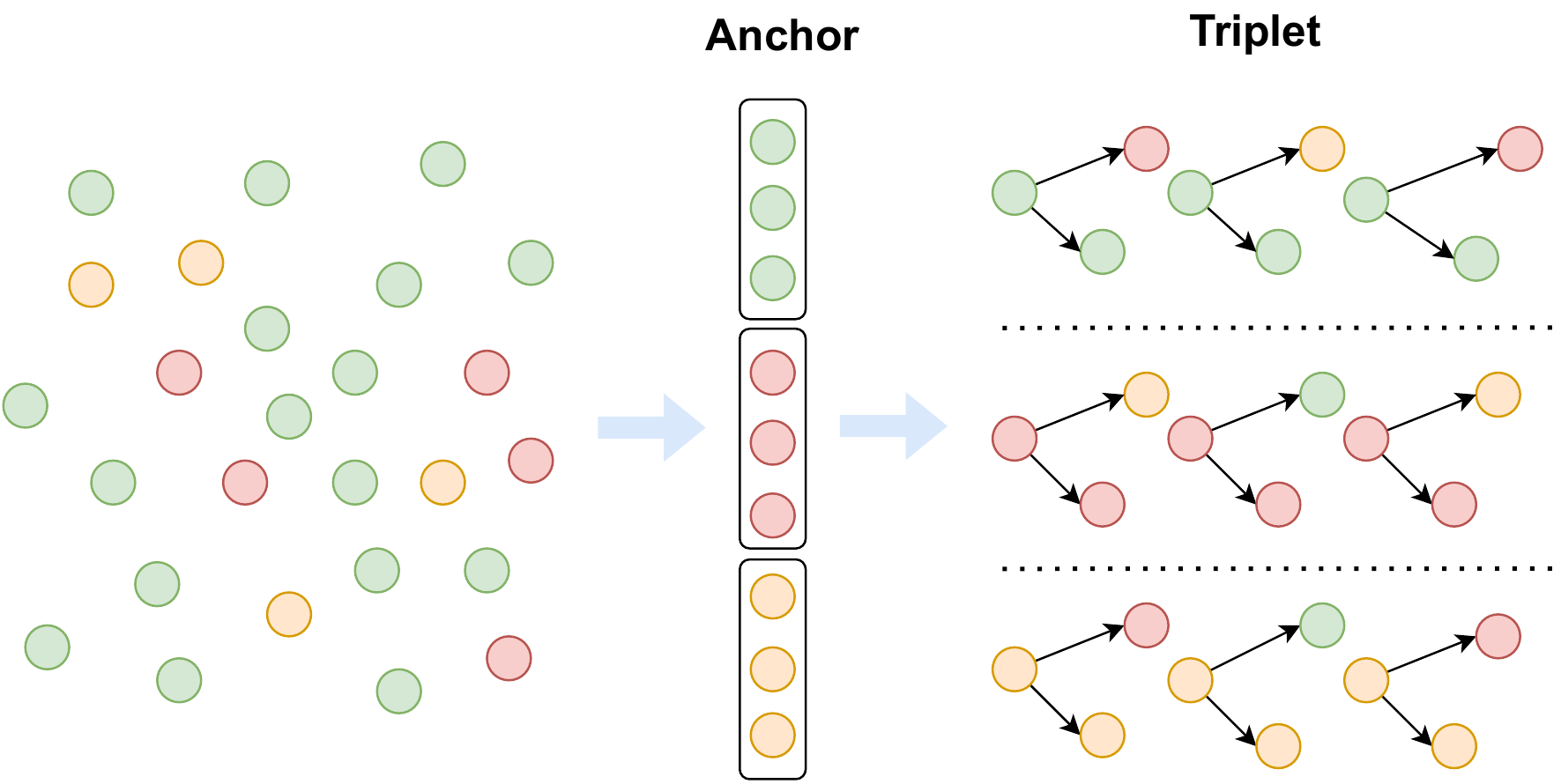}}
 \hspace{0.5in}
 \subfigure[Class-center involved loss]{\includegraphics[width=2in]{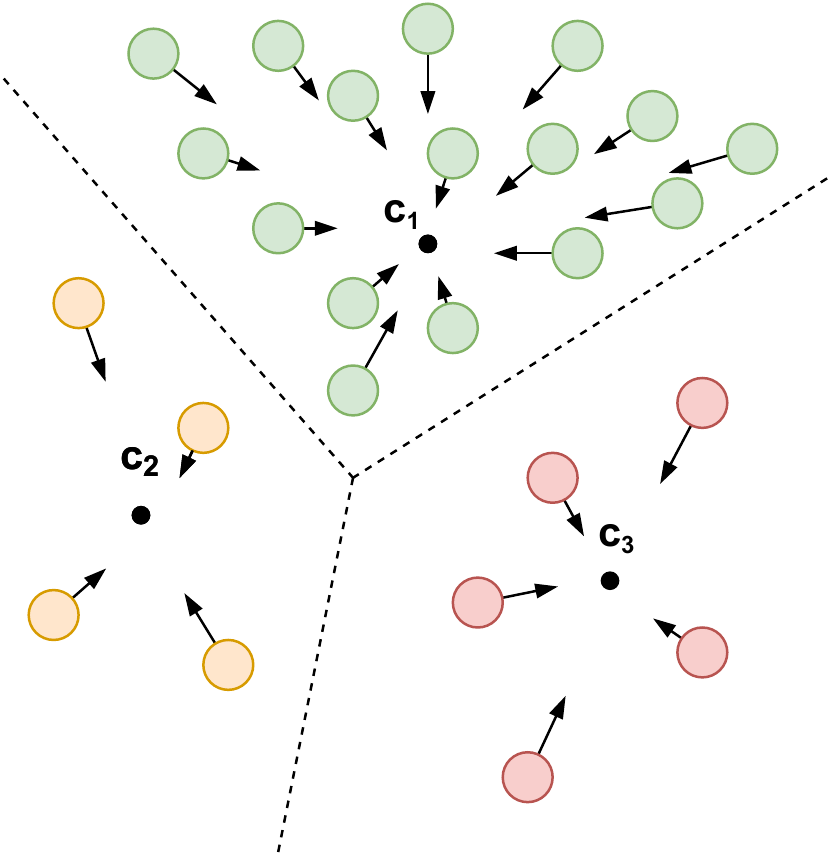}}

        \caption{Demonstration of the proposed method. A two-stage training procedure is proposed to train the feature extractor (a), firstly by the original triplet loss with class-balanced triplet sampler (b), and then by the class-center involved triple loss (c). 
        An original triplet consists of one \textit{anchor} sample, one \textit{positive} sample belonging to the same class of the anchor, and one \textit{negative} sample belonging to another class. Class-center involved triplet consists of one anchor, the class center of the positive, and the class center of the negative. 
        } 
        \label{fig:framework}
    \end{center}
\end{figure*}

In order to accurately diagnose both common and rare diseases, we propose the two-stage Progressive Class-Center Triplet (PCCT) framework, which effectively alleviates the imbalanced issue in medical diagnosis.
The overview of our proposed PCCT is shown in Figure~\ref{fig:framework}(a).
In the first stage, we design a batch-balanced triplet sampler to train the feature extractor with triplet loss. 
The batch-balanced triplet sampler obtains a batch of triplets, whose anchors from different classes are balanced.
Such triplet loss can help alleviate the imbalanced issue during training the feature extractor, and the outputs of the trained feature extractor would help capture a coarse class center for each class. 
In the second stage, we design a class-center involved triplet loss to capture more compact representations based on the coarse class centers from the first stage.
Class-center involved triplet loss replaces the positive and negative samples with their corresponding class centers, which can enforce the feature representation of each data to be closer to the corresponding class center and therefore help the distribution of each class more compact in feature space.
To demonstrate the generality of the proposed method, the two-stage optimization procedure is finally extended in two aspects, i.e., directly learning class centers by considering the class centers as trainable model parameters, and extending to other commonly used losses (e.g., pair-wise and quadruplet losses). 

\subsection{Triplet loss with batch-balanced triplet sampler}


In the first stage, we  propose triplet loss with a batch-balanced triplet sampler to train the feature extractor to capture discriminative features. 
In order to better describe our method, we first briefly introduce  the classical triplet loss. 
Each triplet is composed of an anchor, a positive and a negative, 
where the anchor and the positive are sampled from the same class, and the negative is sampled from other classes. 
Then the triplet loss is adopted to help satisfy the condition that the distance between the anchor and the positive is closer than the distance between the anchor and negative in the feature space, i.e.,
\begin{equation} \label{eq:constraint}
\| \ve{f}(\ve{a}_i; \vv{\theta}) - \ve{f}(\ve{p}_i; \vv{\theta}) \| + \alpha <  
\| \ve{f}(\ve{a}_i; \vv{\theta}) - \ve{f}(\ve{n}_i; \vv{\theta}) \| \,,
\end{equation}
where $\ve{a}_i$, $\ve{p}_i$ and  $\ve{n}_i$ respectively denote an anchor, positive and negative.  $\vv{\theta}$ denotes model parameters of the CNN model (e.g., ResNet-50) to be learned, and $\ve{f}(\cdot; \vv{\theta})$ is the function for the feature extractor part of the CNN model. $\|\cdot\|$ can be any $L_p$ norm ($p=2$ by default), and $\alpha$ is the margin in the inequality constraint~(\ref{eq:constraint}), which is set to 0.5 by default. 
In this way, the triplet loss for the CNN model can be defined as 
\begin{eqnarray} \label{eq:triplet}
l_{t}(\ve{a}_i, \ve{p}_i, \ve{n}_i; \vv{\theta}) =  [ \|\ve{f}(\ve{a}_i; \vv{\theta}) - \ve{f}(\ve{p}_{i};  \vv{\theta}) \| + \alpha  \nonumber \\
 - \|\ve{f}(\ve{a}_i; \vv{\theta})- \ve{f}(\ve{n}_{i}; \vv{\theta}) \| ]_{+} \,,
\end{eqnarray}
with $[ d ]_{+} = \max(0, d)$ denoting the hinge loss. If totally $N$ triplets are involved in model training, the loss function will become $L(\vv{\theta}) = \frac{1}{N}\sum_{i=1}^{N} l_{t}(\ve{a}_i, \ve{p}_i, \ve{n}_i; \vv{\theta}) \,$. 
$L(\vv{\theta})$ can be combined with the cross-entropy loss to optimize the CNN model simultaneously.

In our work, a class-balanced triplet sampler is applied to alleviate the imbalanced issue.
As shown in Figure~\ref{fig:framework}(b), the class-balanced triplet sampler firstly randomly samples an equal number (e.g. 10) of images from each class to form a batch of samples. 
Then, a triplet is formed by assigning a sample from a random class as the anchor and randomly sampling a positive from the same class and a negative from a different class. 
Furthermore, the random-hard triplet sample strategy~\cite{schroff2015facenet} is adopted in the class-balanced triplet sampler to improve training efficiency.
In this way, a batch of class-balanced triplets is formed and used to optimize the triplet loss $L(\vv{\theta})$.
Compared with the previous triplet methods~(e.g., \cite{schroff2015facenet}), our method is more applicable to solve the class imbalance issue. 
The previous triplet losses often ignore class distributions when forming the triplets, causing the triplets in each batch to be more relevant to frequent classes. In contrast, our class-balanced sampler always generates class-balanced batches during model training, i.e., the number of anchors from different classes remains equivalent in each batch in the training process, therefore helping to alleviate the class imbalance issue.
Further, the triplet loss can help capture a coarse class center for each class by enforcing that the distance between the samples from the same class is smaller than that between the samples from different classes.
The coarse class center in the feature space can help the feature extractor in the next stage to capture a more compact feature distribution for each class and help stabilize the training process.
Without the first-stage training process, randomly initialized parameters of the feature extractor would probably cause the distributions to spread more or less randomly and heavily overlap across classes in the feature space especially at the early training stage. Such frequently and largely changed class distribution over training epochs would cause the training instability in the second stage when class centers are involved. 
Therefore, the first stage sets the foundation for the second stage to converge more stably.
%

\subsection{Class-center involved triplet loss}
After initially training the CNN feature extractor model in the first stage, a class-center involved triplet loss is designed to further improve the feature extractor in the second stage.
The new triplet loss would help enforce that the feature representation of each data becomes closer to the corresponding class center in the feature space, therefore making the overall class-level distribution more compact in the feature space.
For the frequent (larger-sample) classes in a large-scale medical dataset, the images within the same class may be  visually quite different, leading to a spreading distribution of this class in a relatively large region in the feature space. As a result, more regions could be simultaneously occupied by multiple classes due to the larger spreading distributions of the larger-sample classes in the feature space, which would cause worse classification performance within such regions.
Ideally, a model trained with triplet loss would help each class have more compact distribution in the feature space. However, due to possible large visual difference between images of the same large-sample class, images from the same large-sample class but belonging to two triplets may cause the triplet inequality constraints to be satisfied for one triplet but not for another over training iterations, which could lead to longer training for the optimization to be convergent.



To further improve the training performance, we propose modifying the original triplet loss with class centers such that global information on distributions of all classes can be considered in the modified triplet loss.
As shown in Figure~\ref{fig:framework}(c), we design the class-center involved triplet loss, i.e.,
\begin{eqnarray} \label{eq:center}
l_{c}(\ve{a}_i, \ve{p}_i, \ve{n}_i; \vv{\theta}_t) = [\|\ve{f}(\ve{a}_i; \vv{\theta}_t) - \ve{c}(\ve{a}_{i}; \vv{\theta}_{t-1}) \| + \alpha \nonumber \\
- \|\ve{f}(\ve{a}_i; \vv{\theta}_t) - \ve{c}(\ve{n}_{i}; \vv{\theta}_{t-1}) \|]_{+}\,.
\end{eqnarray}
Similar to Equation~(\ref{eq:triplet}), $\ve{f}(.;\vv{\theta})$ denotes a CNN feature extractor model with parameters $\vv{\theta}$ to extract feature representation for any input image.
Differently, the extracted feature $\ve{f}(\ve{p}_i; \vv{\theta})$ and $\ve{f}(\ve{n}_i; \vv{\theta})$ of the positive and negative samples in Equation~(\ref{eq:triplet}) are replaced by their corresponding class centers $\ve{c}(\ve{a}_{i}; \vv{\theta}_{t-1})$ and $\ve{c}(\ve{n}_{i}; \vv{\theta}_{t-1})$ respectively.
The class centers  are obtained by averaging the feature representations of all the training samples from the same class of $\ve{p}_i$ and $\ve{n}_i$ respectively, where the features are extracted from the feature extractor trained in the previous epoch.
With a batch of images, each image is used as an anchor and those negative centers contributing non-zero losses in Equation~(\ref{eq:center}) are selected to form triplets with the anchor.
Once the feature extractor is well trained, for each class, the feature representations of all the training samples from the same class are finally averaged to obtain the class center representation. 
The nearest class center method is applied to make a prediction for any new test sample, i.e., classifying the test sample to the class whose class center is closest to the test sample in the feature space. 

Our new triplet loss enforces the samples of each class to be attracted to the class center, such that a more compact distribution is obtained for each class. 
The class-center involved triplet loss of the proposed PCCT method can be viewed as the fusion of the traditional triplet loss~\cite{schroff2015facenet} and the single image-based class-center loss~\cite{wen2016discriminative}. This combination can capture a more compact data distribution of each class in the feature space.
It is also similar to the  triplet-center loss for object retrieval~\cite{he2018triplet}. 
However, our proposed PCCT method uses all negative centers for each {anchor}, while the triplet-center loss from the related work~\cite{he2018triplet} uses only the nearest {negative} center for each {anchor}. That means, the global distribution of all classes in the feature space is considered in our new triplet loss, which may help update the model more efficiently in training. 
Also differently, our class-center triplet loss is proposed particularly to alleviate the class imbalance issue.

\subsection{Extensions of class-center based triplet loss} \label{sec:method_extend}
Our two-stage PCCT can be extended to the pair-wise loss~\cite{chopra2005learning} and the quadruplet loss~\cite{chen2017beyond}, demonstrating the generalization of our class-center involved triplet loss.
Pair-wise loss is one of basic methods in metric learning, and quadruplet loss has been shown to cause larger inter-class variation and smaller intra-class variation in the feature space compared to the triplet loss, which can obtain a better generalization ability in person ReID~\cite{ding2015deep}.
In order to better describe the extended methods, we first  formalize  the commonly used pair-wise loss and quadruplet loss.
The pair-wise ranking loss enforces the paired samples from the same class to be close to each other and the paired samples from different classes to be further apart.
The pair-wise ranking loss $l_p$ is defined with paired samples of same or different classes, 
\begin{eqnarray} \label{eq:double}
l_{p}(\ve{a}_i, \ve{b}_i; \vv{\theta}) = \mathbbm{1}(\ve{a}_i, \ve{b}_i)\|\ve{f}(\ve{a}_i; \vv{\theta}) - \ve{f}(\ve{b}_{i}; \vv{\theta}) \| \nonumber\\ + (1 - \mathbbm{1}(\ve{a}_i, \ve{b}_i))[\alpha -  \|\ve{f}(\ve{a}_i; \vv{\theta}) - \ve{f}(\ve{b}_{i}; \vv{\theta}) \|]_{+} \,,
\end{eqnarray}
where the indicator function $\mathbbm{1}(\ve{a}_i,\ve{b}_i)$ is 1 if $\ve{a}_i$ and $\ve{b}_i$ belong to the same class, and 0 otherwise. 

Compared to triplet loss, the quadruplet loss additionally enforces that the distance between two samples of the same class is smaller than that between two samples from another two classes. 
The quadruplet loss is defined based on quadruple samples which include one anchor and one positive from the same class, and two negative samples ($\ve{n}_{i, 1}$ and $\ve{n}_{i, 2}$) coming from  two other different classes,
\begin{eqnarray}
\label{eq:quadruple}
l_{q}(\ve{a}_i, \ve{p}_i, \ve{n}_{i, 1}, \ve{n}_{i, 2}; \vv{\theta}) =  [ \|\ve{f}(\ve{a}_i; \vv{\theta}) - \ve{f}(\ve{p}_{i};  \vv{\theta}) \| + \alpha  \nonumber \\
 - \|\ve{f}(\ve{a}_i; \vv{\theta})- \ve{f}(\ve{n}_{i, 1}; \vv{\theta}) \| ]_{+} \nonumber \\
 + [ \|\ve{f}(\ve{a}_i; \vv{\theta}) - \ve{f}(\ve{p}_{i};  \vv{\theta}) \| + \beta  \nonumber \\
 - \|\ve{f}(\ve{n}_{i, 1}; \vv{\theta})- \ve{f}(\ve{n}_{i, 2}; \vv{\theta}) \| ]_{+}\,,
\end{eqnarray}
where $\beta$ is another constant that is normally smaller than $\alpha$.

For extension,  the class-center based pair-wise ranking loss and quadruplet loss can be obtained by simply replacing positive and negative samples with the corresponding class centers, which is similar to the class-center based triplet loss in Equation~(\ref{eq:center}), i.e., 
\begin{eqnarray} \label{eq:double_center}
l_{pc}(\ve{a}_i, \ve{b}_i; \vv{\theta}_t) = \mathbbm{1}(\ve{a}_i, \ve{b}_i) \|\ve{f}(\ve{a}_i; \vv{\theta}_{t}) - \ve{c}(\ve{b}_{i}; \vv{\theta}_{t-1}) \| \nonumber\\ +  (1 - \mathbbm{1}(\ve{a}_i, \ve{b}_i)) [\alpha -  \|\ve{f}(\ve{a}_i; \vv{\theta}_{t}) - \ve{c}(\ve{b}_{i}; \vv{\theta}_{t-1}) \|]_{+} \,,
\end{eqnarray}
\begin{eqnarray}
\label{eq:quadruple_center}
l_{qc}(\ve{a}_i, \ve{p}_i, \ve{n}_{i,1}, \ve{n}_{i,2}; \vv{\theta}_{t}) =  [ \|\ve{f}(\ve{a}_i; \vv{\theta}_{t}) - \ve{c}(\ve{p}_{i}; \vv{\theta}_{t-1}) \| + \alpha  \nonumber \\
 - \|\ve{f}(\ve{a}_i; \vv{\theta}_{t})- \ve{c}(\ve{n}_{i,1}; \vv{\theta}_{t-1}) \| ]_{+} \nonumber \\
 + [ \|\ve{f}(\ve{a}_i; \vv{\theta}_{t}) - \ve{c}(\ve{p}_{i};  \vv{\theta}_{t-1}) \| + \beta  \nonumber \\
 - \|\ve{c}(\ve{n}_{i,1}; \vv{\theta}_{t-1})- \ve{c}(\ve{n}_{i,2}; \vv{\theta}_{t-1}) \| ]_{+} \,.
\end{eqnarray}
Once the feature extractor is well trained by the pair-wise ranking loss or the quadruplet loss, it can be used for classification of any test sample using the nearest class-center method as mentioned above, which is easily accessible in the testing stage.

\subsection{Trainable class centers}
Besides the extension to other losses, another extension is to consider the class centers as a trainable model parameter. 
In the class-center involved triplet loss, each class center is estimated by averaging the feature representations of all training samples from the same class based on the temporary feature extractor trained at the previous training epoch. Such an estimate is time-consuming and needs to be performed at each epoch. One way to reduce the computational cost during feature extractor training is to consider each class center as part of the model parameters which will be optimized with the feature extractor simultaneously. This can be easily realized by replacing the estimated class center (i.e., $\ve{c}(\cdot; \vv{\theta}_{t-1})$) in Equation~(\ref{eq:center}) with corresponding trainable class centers. 
Formally, we define the class centers $\ve{C} = \{\ve{c}_1, \ve{c}_2,..., \ve{c}_K\}$ as part of model parameters, where $K$ represents the total number of classes. With the gradient back-propagated from the loss function (e.g., Equation~\ref{eq:center}), the trainable class centers $\ve{c}$ can be easily updated together with parameters of the feature extractor $\ve{f}(\cdot; \vv{\theta})$, instead of going over the training data to obtain the averaging feature representation.
Due to the limited number of parameters for the representation of the trainable class centers, the  additional computational cost caused by the trainable class centers during training is negligible. 
Specially, we proposed to introduce this strategy into the original PCCT and called the new version Efficient-PCCT. 

A recently study introduces trainable class centers~\cite{Hayat2019GaussianAF} to maximize a Gaussian affinity objective with hyper-parameters to be tuned and approximately equal distance between class centers to be enforced. Differently, our trainable class centers can be simply obtained by minimizing the proposed class-center involved triplet loss without extra constraints.


 \section{Experimental Evaluation}
 \label{sec:experiments}

	\begin{table}[!t]
	    \centering
	    \caption{The statistics of three medical image datasets.}
	    \scalebox{0.95}{
	    \begin{tabular}{c|ccc}
	        \hline
	        Dataset & Class number & \begin{tabular}[c]{@{}c@{}}Image number\\ in largest class\end{tabular} &\begin{tabular}[c]{@{}c@{}}Image number\\ in smallest class\end{tabular} \\ \hline
	        Skin7~\cite{tschandl2018ham10000,codella2019skin} & 7 & 6705 & 115\\ 
	        Skin198~\cite{sun2016benchmark} & 198 & 60 & 10\\ 
	        ChestXray-COVID~\cite{cohen2020covid} & 3 & 8851 & 133 \\
	        \hline
	    \end{tabular}
	    }
	    \label{tab:datasets}
	\end{table}

\begin{table}[!t]
    \centering
    \caption{The batch size during training for each dataset.}
    \begin{tabular}{c|ccc}
        \hline
        ~ & Skin7 & Skin198 & ChestXray-COVID\\ 
        \hline
        First stage & 10 per class & 5 per class & 5 per class\\ 
        Second stage & 16 & 32 & 8\\ 
  
        \hline
    \end{tabular}
    \label{tab:batch}
\end{table}
	
\begin{table*}[!ht]
    \centering
    \caption{Performance of the proposed approach and baseline methods on Skin7, Skin198 and ChestXray-covid.}
    \label{tab:OverallPerformance}
    {
        \begin{tabular}{c|ccc|ccc|ccc}
    \toprule
    
          \multirow{2}{*}{Methods}  & \multicolumn{3}{c}{Skin 7} & \multicolumn{3}{c}{Skin 198}& \multicolumn{3}{c}{ChestXray-COVID} \\
         \cmidrule(r){2-4}\cmidrule(r){5-7}\cmidrule(r){8-10}
         
          & MF1 & MCP &\multicolumn{1}{c}{MCR} & MF1 & MCP &\multicolumn{1}{c}{MCR} & MF1 & MCP &\multicolumn{1}{c}{MCR} \\
         \midrule
         BCE &\tabincell{c}{  83.65 \\ \scriptsize{(1.52)} } & \tabincell{c}{86.96 \\ \scriptsize{(1.96)}} & \tabincell{c}{81.15 \\ \scriptsize{(1.62)}} & \tabincell{c}{51.91 \\ \scriptsize{(1.10)}} & \tabincell{c}{56.41 \\ \scriptsize{(1.27)}}  & \tabincell{c}{52.12 \\ \scriptsize{(1.14)}}  & \tabincell{c}{82.37 \\ \scriptsize{(2.97)}} & \tabincell{c}{91.35 \\ \scriptsize{(3.17)}} & \tabincell{c}{76.95 \\ \scriptsize{(3.35)}} \\
         WCE & \tabincell{c}{82.45 \\ \scriptsize{(1.31)}} & \tabincell{c}{83.35 \\ \scriptsize{(1.79)}} & \tabincell{c}{82.06 \\ \scriptsize{(1.47)} }& \tabincell{c}{60.21 \\ \scriptsize{(1.36)}} &  \tabincell{c}{64.82 \\ \scriptsize{(1.34)}} & \tabincell{c}{60.23 \\ \scriptsize{(1.12)}} & \tabincell{c}{82.18 \\ \scriptsize{(5.44)}} & \tabincell{c}{ 93.59\\ \scriptsize{(1.99)}} & \tabincell{c}{76.24 \\ \scriptsize{(6.71)}}  \\
         OCE & \tabincell{c}{83.53 \\ \scriptsize{(1.33)}} & \tabincell{c}{87.26 \\ \scriptsize{(1.27)}} & \tabincell{c}{80.81 \\ \scriptsize{(1.39)}} & \tabincell{c}{59.77 \\ \scriptsize{(1.89)}} & \tabincell{c}{64.87 \\ \scriptsize{(2.06)}} & \tabincell{c}{59.34 \\ \scriptsize{(1.87)}} & \tabincell{c}{84.16 \\ \scriptsize{(4.51)}} & \tabincell{c}{92.21 \\ \scriptsize{(3.55)}}& \tabincell{c}{78.77 \\ \scriptsize{(5.11)}} \\
         WFCE & \tabincell{c}{83.52 \\ \scriptsize{(1.63)}} & \tabincell{c}{86.43 \\ \scriptsize{(1.34)}} & \tabincell{c}{81.25 \\ \scriptsize{(1.78)}} & \tabincell{c}{53.28 \\ \scriptsize{(2.65)}} & \tabincell{c}{58.31 \\ \scriptsize{(2.77)}} & \tabincell{c}{53.34 \\ \scriptsize{(2.58)}} & \tabincell{c}{83.77 \\ \scriptsize{(3.71)}} & \tabincell{c}{91.31 \\ \scriptsize{(3.08)}} & \tabincell{c}{79.32 \\ \scriptsize{(4.15)}}\\
         
         
         TSD & \tabincell{c}{86.00 \\ \scriptsize{(1.02)}} & \tabincell{c}{87.74 \\ \scriptsize{(1.31)}} & \tabincell{c}{84.63 \\ \scriptsize{(1.19)}} & \tabincell{c}{64.23 \\ \scriptsize{(1.54)}} & \tabincell{c}{67.10 \\ \scriptsize{(1.90)}} & \tabincell{c}{65.62 \\ \scriptsize{(1.56)}} & \tabincell{c}{91.13 \\ \scriptsize{(0.76)}} & \tabincell{c}{94.22 \\ \scriptsize{(1.69)}} & \tabincell{c}{88.47 \\ \scriptsize{(1.28)}}\\
         \midrule
         
         PCCT & \tabincell{c}{86.20 \\ \scriptsize{(1.07)}} & \tabincell{c}{87.77 \\ \scriptsize{(1.54)} } & \tabincell{c}{\textbf{84.98} \\ \scriptsize{(0.75)}} &
         \tabincell{c}{\textbf{65.20} \\ \scriptsize{(1.49)}} & \tabincell{c}{68.40 \\ \scriptsize{(1.36)}} & \tabincell{c}{\textbf{66.02} \\ \scriptsize{(1.50)}} &
         \tabincell{c}{\textbf{91.32} \\ \scriptsize{(0.73)}} & \tabincell{c}{\textbf{94.67} \\ \scriptsize{(0.66)}} &\tabincell{c}{\textbf{88.49}\\ \scriptsize{(1.35)}} \\
         
         Efficient-PCCT & \tabincell{c}{{\textbf{86.36}} \\ \scriptsize{(0.95)}} & \tabincell{c}{{\textbf{88.97}} \\ \scriptsize{(1.51)}} & \tabincell{c}{{84.32} \\ \scriptsize{(0.70)}} & \tabincell{c}{{64.51} \\ \scriptsize{(1.65)}} &  \tabincell{c}{{\textbf{68.76}} \\ \scriptsize{(1.80)}} & \tabincell{c}{{64.74} \\ \scriptsize{(1.84)}}  & \tabincell{c}{{89.91} \\ \scriptsize{(1.44)}} & \tabincell{c}{{93.57} \\ \scriptsize{(2.42)} } & \tabincell{c}{86.95 \\ \scriptsize{(2.59)}} \\
         
         \bottomrule
    \end{tabular}
    }
\end{table*}

\begin{table*}[!ht]
    \centering
    \caption{Performance on small-sample classes on Skin7, Skin198 and ChestXray-covid.}
    \label{tab:SmallClasses_Performance}
    {
        \begin{tabular}{c|ccc|ccc|ccc}
    \toprule
          \multirow{2}{*}{Methods}   & \multicolumn{3}{c}{Skin 7} & \multicolumn{3}{c}{Skin 198}& \multicolumn{3}{c}{ChestXray-COVID} \\
         \cmidrule(r){2-4}\cmidrule(r){5-7}\cmidrule(r){8-10}
         \multicolumn{1}{c|}{ }
          & MF1 & MCP &\multicolumn{1}{c}{MCR} & MF1 & MCP &\multicolumn{1}{c}{MCR} & MF1 & MCP &\multicolumn{1}{c}{MCR} \\
         \midrule
         BCE &\tabincell{c}{  73.67 \\ \scriptsize{(3.62)} } & \tabincell{c}{79.03 \\ \scriptsize{(0.76)}} & \tabincell{c}{69.39 \\ \scriptsize{(6.24)}} & \tabincell{c}{18.59 \\ \scriptsize{(2.43)}} & \tabincell{c}{24.22 \\ \scriptsize{(3.00)}}  & \tabincell{c}{16.67 \\ \scriptsize{(2.78)}}  & \tabincell{c}{ 71.24 \\ \scriptsize{(6.65)}} & \tabincell{c}{ 95.11\\ \scriptsize{(5.02)}} & \tabincell{c}{ 57.26\\ \scriptsize{(7.86)}} \\
         WCE & \tabincell{c}{77.96 \\ \scriptsize{(5.31)}} & \tabincell{c}{87.18 \\ \scriptsize{(2.47)}} & \tabincell{c}{70.83 \\ \scriptsize{(7.64)} }& \tabincell{c}{53.37 \\ \scriptsize{(1.99)}} &  \tabincell{c}{65.21 \\ \scriptsize{(2.52)}} & \tabincell{c}{49.79 \\ \scriptsize{(2.68)}} & \tabincell{c}{69.85 \\ \scriptsize{(16.84)}} & \tabincell{c}{ \textbf{97.60}\\ \scriptsize{(5.37)}} & \tabincell{c}{ 57.15\\ \scriptsize{20.19}}  \\
         OCE & \tabincell{c}{74.05 \\ \scriptsize{(8.91)}} & \tabincell{c}{84.93 \\ \scriptsize{(5.16)}} & \tabincell{c}{66.17 \\ \scriptsize{(11.81)}} & \tabincell{c}{56.41 \\ \scriptsize{(3.55)}} & \tabincell{c}{66.46 \\ \scriptsize{(4.25)}} & \tabincell{c}{53.42 \\ \scriptsize{(3.17)}} & \tabincell{c}{ 72.72\\ \scriptsize{(13.09)}} & \tabincell{c}{ 91.52\\ \scriptsize{(9.33)}} & \tabincell{c}{ 61.00\\ \scriptsize{(14.90)}} \\
         WFCE & \tabincell{c}{76.21 \\ \scriptsize{(4.94)}} & \tabincell{c}{84.96 \\ \scriptsize{(3.61)}} & \tabincell{c}{69.35 \\ \scriptsize{(6.62)}} & \tabincell{c}{20.36 \\ \scriptsize{(2.08)}} & \tabincell{c}{26.83 \\ \scriptsize{(2.74)}} & \tabincell{c}{17.99 \\ \scriptsize{(2.21)}} & \tabincell{c}{ 72.52\\ \scriptsize{(10.91)}} & \tabincell{c}{ 93.64\\ \scriptsize{(7.51)}} & \tabincell{c}{ 60.31\\ \scriptsize{(13.58)}}\\
         
         
        TSD & \tabincell{c}{80.73 \\ \scriptsize{(7.76)}} & \tabincell{c}{87.11 \\ \scriptsize{(7.51)}} & \tabincell{c}{75.65 \\ \scriptsize{(9.76)}} & \tabincell{c}{62.89 \\ \scriptsize{(3.57)}} & \tabincell{c}{66.65 \\ \scriptsize{(4.12)}} & \tabincell{c}{65.10 \\ \scriptsize{(3.67)}} & \tabincell{c}{\textbf{83.47} \\ \scriptsize{(1.69)}} & \tabincell{c}{88.13 \\ \scriptsize{(2.37)}} & \tabincell{c}{\textbf{79.40} \\ \scriptsize{(3.26)}}\\
         
         \midrule
         
         PCCT & \tabincell{c}{\textbf{81.40} \\ \scriptsize{(5.48)}} & \tabincell{c}{84.16 \\ \scriptsize{(2.67)} } & \tabincell{c}{\textbf{79.13} \\ \scriptsize{(8.43)}} & \tabincell{c}{63.87 \\ \scriptsize{(3.21)}} & \tabincell{c}{67.43 \\ \scriptsize{(2.74)}} & \tabincell{c}{\textbf{65.33} \\ \scriptsize{(3.88)}} & \tabincell{c}{82.62 \\ \scriptsize{(1.52)}} & \tabincell{c}{88.11 \\ \scriptsize{(1.89)}} & \tabincell{c}{ 77.90 \\ \scriptsize{(3.28)}}  \\
         
          Efficient-PCCT & \tabincell{c}{{{81.11}} \\ \scriptsize{(5.59)}} & \tabincell{c}{{\textbf{90.32}} \\ \scriptsize{(2.42)}} & \tabincell{c}{{73.91} \\ \scriptsize{(8.25)}} & \tabincell{c}{\textbf{64.93} \\ \scriptsize{(3.64)}} &  \tabincell{c}{{\textbf{71.44}} \\ \scriptsize{(3.43)}} & \tabincell{c}{{64.21} \\ \scriptsize{(4.04)}}  & \tabincell{c}{{81.97} \\ \scriptsize{(1.59)}} & \tabincell{c}{{88.77} \\ \scriptsize{(2.19)} } & \tabincell{c}{76.20 \\ \scriptsize{(2.40)}} \\
         
         \bottomrule
    \end{tabular}
    }
\end{table*}

\subsection{Experiment settings}
We choose three challenging datasets (Table~\ref{tab:datasets}) for imbalanced medical diagnosis to evaluate the proposed PCCT method.
All three datasets include frequent and infrequent classes, with varying levels of class imbalance.
The dermoscopy dataset Skin7 contains 7 categories with an imbalance ratio of 58.3 (i.e., 6705/115). Another skin image dataset Skin198 contains 198 categories with an imbalance ratio of 6.0. The X-ray dataset ChestXray-COVID contains 3 categories with an imbalance ratio of 66.5, including 8,851 ``Normal'' images, 1,000 ``Pneumonia'' images and 133 ``COVID'' images. 
During training, all Skin7 and Skin198 images are resized to $300\times 300$ pixels and then randomly cropped to $224\times 224$ pixels, while ChestXray-COVID images are resized to 512 followed by a random horizontal flip. 
The batch size setting (Table~\ref{tab:batch}) is chosen for better convergence.

For evaluation, the proposed PCCT is compared with multiple relevant baselines and adopts the same backbone (ResNet-50 by default) with these baselines. For our PCCT, the last  output layer of the original CNN is replaced by a new fully-connected layer whose output is a 128-dimensional feature vector.
Adam optimizer is adopted for model training, with learning rate $0.0001$, $\beta_1=0.9$, and $\beta_2=0.99$. $\alpha$ is set to 0.5 for training the feature extractor. Our proposed PCCT is evaluated with 5-fold cross-validation.
Three evaluation metrics are adopted to obtain comprehensive performance comparison, including average precision (`MCP'), average recall (`MCR'), and average F1 score (`MF1') over classes.
Both stages are trained for 200 epochs to ensure the convergence of the CNN model. Specially, MF1 takes both MCP and MCR into account, which is considered as a more comprehensive measurement to evaluate the model. Note that all the measurements are reported in the form of percentage (i.e., \%), which is omitted in brief.
The means and standard deviations (in bracket) of the three measurements over five folds are reported in each evaluation.


\subsection{Effectiveness of the triplet-based approach} \label{comparison_skin_datasets}
	
	%

The effectiveness of the proposed PCCT is evaluated by comparing with multiple widely-used training strategies for the class imbalance issue, including  the cross-entropy loss with class re-weighting (`WCE'), the cross-entropy loss with the strategy of oversampling (`OCE'), the focal loss with class re-weighting (`WFCE'), and the state-of-the-art two-stage decoupling method (`TSD')~\cite{Kang2020DecouplingRA}. The basic cross-entropy loss (`CE') without any class re-balancing strategy is also included for comparison. In training, the batch size is set to 32 for all baselines and the same training and evaluation protocols are used as for the proposed approach.
Table~\ref{tab:OverallPerformance} shows that the proposed PCCT performs best on all the datasets. For example, on the Skin198 dataset, PCCT achieves the mean F1 score of 86.2\%, which is clearly better than all the other methods.
Furthermore, the extensive version (Efficient-PCCT) which directly represents the class centers by additional trainable model parameters is also compared in Table~\ref{tab:OverallPerformance}. Initial evaluation shows that the triplet loss with trainable class centers (Efficient-PCCT) performs similarly well compared to the original PCCT with computed class centers on all three datasets. Note that the original PCCT needs all the training dataset to obtain the class centers at each training epoch, while the Efficient-PCCT simply updates the class center along with the model parameters. 
Considering its feasibility and effectiveness, trainable class centers may be a better choice than the estimated class centers computed at each training epoch especially when the training dataset is large.

To demonstrate the effect of the proposed PCCT on infrequent classes, we also compared the performance of all methods on the most infrequent (small-sample) class from Skin7 (115 images) and ChestXray-COVID (133 images), and the average performance on 70 most infrequent classes (20 or fewer images per class) from Skin198. Table~\ref{tab:SmallClasses_Performance} shows that on all three datasets, the proposed PCCT achieves sharp improvement on the small-sample classes, e.g., the improvement is about $7.35\%$ (mean F1 score) on the small class of Skin7 from the baseline OCE to the proposed PCCT (Table~\ref{tab:SmallClasses_Performance}).
Meanwhile, the mean F1 performance over all classes also obtains a $2.7\%$ improvement (Table~\ref{tab:OverallPerformance}).
Similar improvement on infrequent classes is observed on Skin198 and ChestXray-COVID. 
Compared with the baseline OCE, our proposed PCCT achieves $7.46\%$ and $9.2\%$ improvement on infrequent classes of Skin198 and ChestXray-COVID, respectively.  
Wilcoxon Rank test demonstrates that the proposed PCCT and the state-of-the-art approach TSD are significantly better than all the other strong baselines (p-values $<$ 0.05), while there is no significant difference between PCCT and TSD.

\begin{table*}[!ht]
	\begin{center}
	\caption{Classification performance with various model architectures on Skin198.} \label{tab:architectures}
    {
	\begin{tabular}{c|ccc|ccc|ccc|ccc}
    \toprule
           \multirow{2}{*}{Methods}  & \multicolumn{3}{c}{ResNet-50} & \multicolumn{3}{c}{DenseNet-121}& \multicolumn{3}{c}{Inception-v4} & \multicolumn{3}{c}{VGG-19}\\
         \cmidrule(r){2-4}\cmidrule(r){5-7}\cmidrule(r){8-10}\cmidrule(r){11-13}
         \multicolumn{1}{c|}{ }
          & MF1 & MCP &\multicolumn{1}{c}{MCR} & MF1 & MCP &\multicolumn{1}{c}{MCR} & MF1 & MCP &\multicolumn{1}{c}{MCR} & MF1 & MCP &\multicolumn{1}{c}{MCR}\\
          \midrule
				
			BCE   & \tabincell{c}{  51.91 \\ \scriptsize{(1.10)} }& \tabincell{c}{  56.41 \\ \scriptsize{(1.27)} }& \tabincell{c}{  52.12 \\ \scriptsize{(1.14)} }& \tabincell{c}{  62.91 \\ \scriptsize{(1.37)} }& \tabincell{c}{ 65.95 \\ \scriptsize{(1.42)} }& \tabincell{c}{ 64.00 \\ \scriptsize{(1.55)} }& \tabincell{c}{  50.22 \\ \scriptsize{(1.78)} }& \tabincell{c}{  53.73 \\ \scriptsize{(2.00)} }& \tabincell{c}{  50.83 \\ \scriptsize{(1.79)} } & \tabincell{c}{  51.16 \\ \scriptsize{(1.82)} }& \tabincell{c}{  53.81 \\ \scriptsize{(2.07)} }& \tabincell{c}{  52.51 \\ \scriptsize{(1.85)} }\\ 

			WCE   & \tabincell{c}{  60.21 \\ \scriptsize{(1.36)} }& \tabincell{c}{  64.82 \\ \scriptsize{(1.34)} }& \tabincell{c}{  60.23 \\ \scriptsize{(1.12)} }& \tabincell{c}{  55.04 \\ \scriptsize{(1.74)} }& \tabincell{c}{  61.42 \\ \scriptsize{(2.01)} }& \tabincell{c}{  54.62 \\ \scriptsize{(1.68)} }& \tabincell{c}{  57.92 \\ \scriptsize{(1.71)} }& \tabincell{c}{  62.30 \\ \scriptsize{(1.24)} }& \tabincell{c}{  58.08 \\ \scriptsize{(2.05)} }& \tabincell{c}{  47.09 \\ \scriptsize{(7.91)} }& \tabincell{c}{  50.43 \\ \scriptsize{(7.71)} }& \tabincell{c}{  48.08 \\ \scriptsize{(7.67)} }\\ 

			OCE   & \tabincell{c}{  59.77 \\ \scriptsize{(1.89)} }& \tabincell{c}{  64.87 \\ \scriptsize{(2.06)} }& \tabincell{c}{  59.34 \\ \scriptsize{(1.87)} }& \tabincell{c}{  57.72 \\ \scriptsize{(1.91)} }& \tabincell{c}{  63.86 \\ \scriptsize{(1.85)} }& \tabincell{c}{  56.96 \\ \scriptsize{(1.92)} }& \tabincell{c}{  56.79 \\ \scriptsize{(2.51)} }& \tabincell{c}{  61.83 \\ \scriptsize{(2.70)} }& \tabincell{c}{  56.49 \\ \scriptsize{(2.46)} } & \tabincell{c}{  50.85 \\ \scriptsize{(1.42)} }& \tabincell{c}{  53.99 \\ \scriptsize{(1.44)} }& \tabincell{c}{  52.09 \\ \scriptsize{(1.67)} }\\ 


			WFCE   & \tabincell{c}{  53.28 \\ \scriptsize{(2.65)} }& \tabincell{c}{  58.31 \\ \scriptsize{(2.77)} }& \tabincell{c}{  53.34 \\ \scriptsize{(2.58)} }& \tabincell{c}{  43.03 \\ \scriptsize{(1.28)} }& \tabincell{c}{  46.70 \\ \scriptsize{(1.08)} }& \tabincell{c}{  44.00 \\ \scriptsize{(1.37)} }& \tabincell{c}{  49.88 \\ \scriptsize{(2.65)} }& \tabincell{c}{  53.53 \\ \scriptsize{(2.33)} }& \tabincell{c}{  50.44 \\ \scriptsize{(2.86)} } & \tabincell{c}{  37.13 \\ \scriptsize{(1.98)} } & \tabincell{c}{  39.68 \\ \scriptsize{(2.41)} } & \tabincell{c}{  38.61 \\ \scriptsize{(1.73)} }\\ 
			
			
			TSD & \tabincell{c}{{{64.23}} \\ \scriptsize{(1.54)}} & \tabincell{c}{{{67.10}} \\ \scriptsize{(1.90)}} & \tabincell{c}{{65.62} \\ \scriptsize{(1.56)}} & \tabincell{c}{{62.94} \\ \scriptsize{(1.54)}} &  \tabincell{c}{{{65.99}} \\ \scriptsize{(1.47)}} & \tabincell{c}{{64.04} \\ \scriptsize{(1.71)}}  & \tabincell{c}{\textbf{59.49} \\ \scriptsize{(1.90)}} & \tabincell{c}{{62.13} \\ \scriptsize{(2.14)} } & \tabincell{c}{\textbf{61.01} \\ \scriptsize{(2.06)}} & \tabincell{c}{50.61 \\ \scriptsize{(1.93)}} & \tabincell{c}{53.99 \\ \scriptsize{(1.88)}} & \tabincell{c}{51.66 \\ \scriptsize{(2.06)}} \\
			
	    \midrule



			PCCT  & \tabincell{c}{\textbf{65.20} \\ \scriptsize{(1.49)}} & \tabincell{c}{68.40 \\ \scriptsize{(1.36)}} & \tabincell{c}{\textbf{66.02} \\ \scriptsize{(1.50)}}& \tabincell{c}{  {62.62} \\ \scriptsize{(2.18)} }& \tabincell{c}{  {64.94} \\ \scriptsize{(2.41)} }& \tabincell{c}{  {64.26} \\ \scriptsize{(1.98)} }& \tabincell{c}{ {59.25} \\ \scriptsize{(1.18)} }& \tabincell{c}{  {62.43} \\ \scriptsize{(1.48)} }& \tabincell{c}{  {60.12} \\ \scriptsize{(1.12)} } & \tabincell{c}{   \textbf{52.49} \\ \scriptsize{(2.33)} }& \tabincell{c}{   \textbf{55.39} \\ \scriptsize{(2.84)} }& \tabincell{c}{   \textbf{53.49} \\ \scriptsize{(2.50)} } \\
			 
			 Efficient-PCCT & \tabincell{c}{{{64.51}} \\ \scriptsize{(1.65)}} & \tabincell{c}{\textbf{68.76} \\ \scriptsize{(1.80)}} & \tabincell{c}{{64.74} \\ \scriptsize{(1.84)}} & \tabincell{c}{\textbf{63.43} \\ \scriptsize{(2.01)}} &  \tabincell{c}{{\textbf{66.70}} \\ \scriptsize{(2.14)}} & \tabincell{c}{\textbf{64.36} \\ \scriptsize{(1.98)}}  & \tabincell{c}{{58.74} \\ \scriptsize{(1.43)}} & \tabincell{c}{\textbf{63.12} \\ \scriptsize{(2.11)} } & \tabincell{c}{60.03 \\ \scriptsize{(1.13)}} & \tabincell{c}{42.63 \\ \scriptsize{(2.10)}} & \tabincell{c}{46.09 \\ \scriptsize{(3.15)}} & \tabincell{c}{44.38 \\ \scriptsize{(2.00)}} \\
			 
			\bottomrule
			\end{tabular}
		}
	\end{center}
\end{table*}
	


\subsection{Generalizability and hyper-parameter evaluation}
\begin{figure}[!th]
    \begin{center}
      \includegraphics[width=0.7\linewidth]{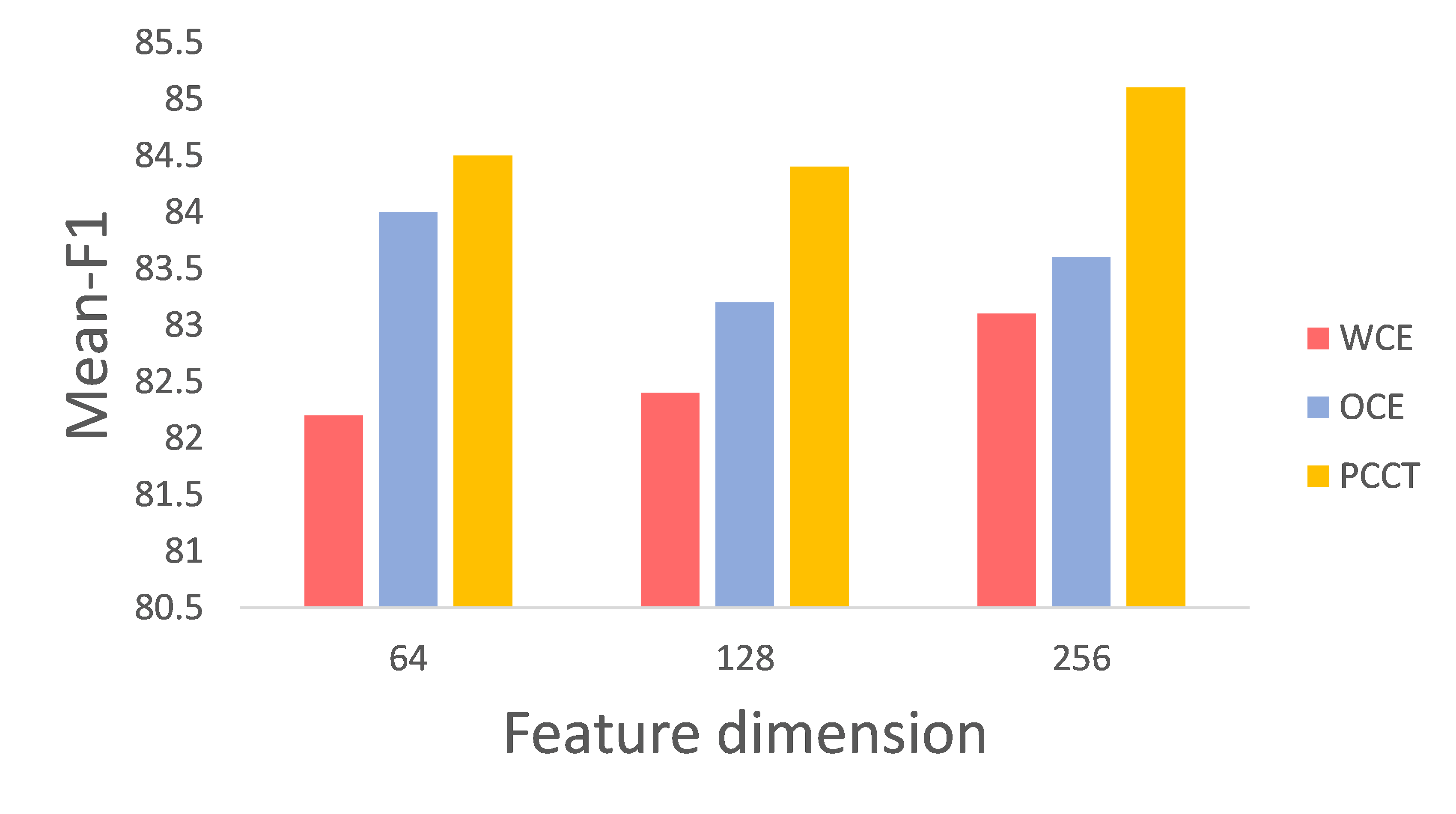}
        \caption{Effect of output dimension on imbalanced classification. Skin7 is used here.} 
        \label{fig:dimension}
    \end{center}
\end{figure}

\begin{figure*}[!ht]
    \begin{center}
       \includegraphics[width=0.3\linewidth]{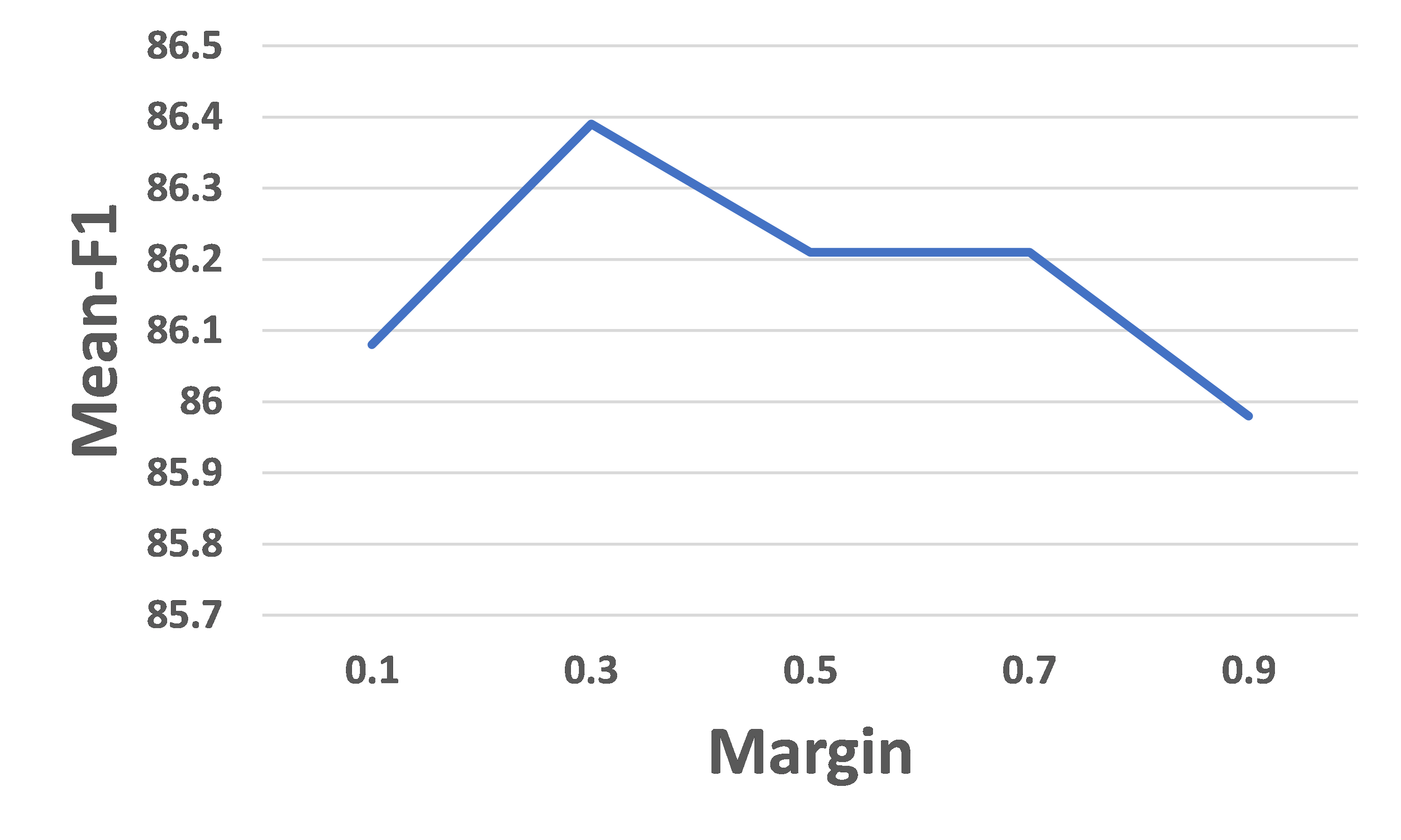}
       \includegraphics[width=0.3\linewidth]{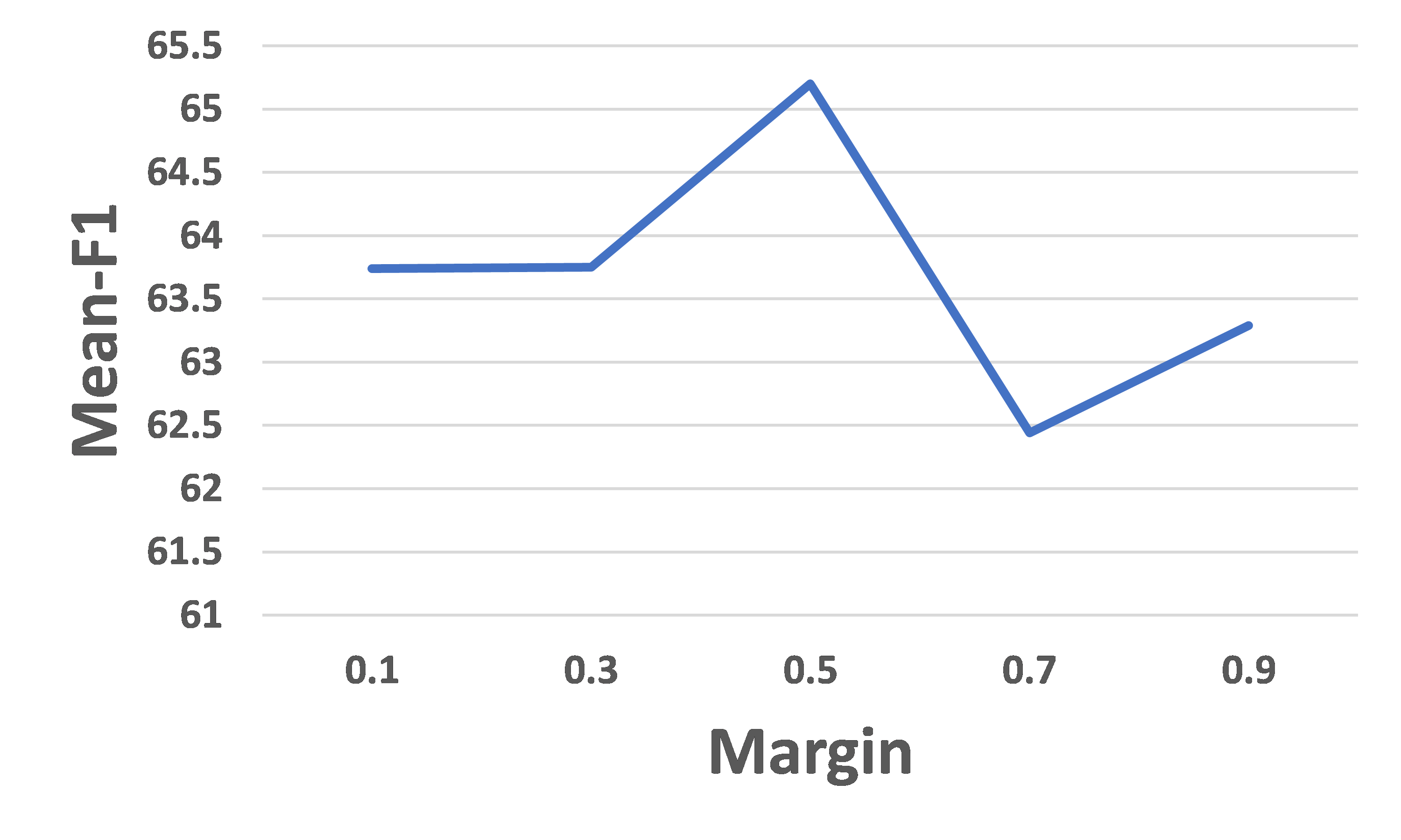}
       \includegraphics[width=0.3\linewidth]{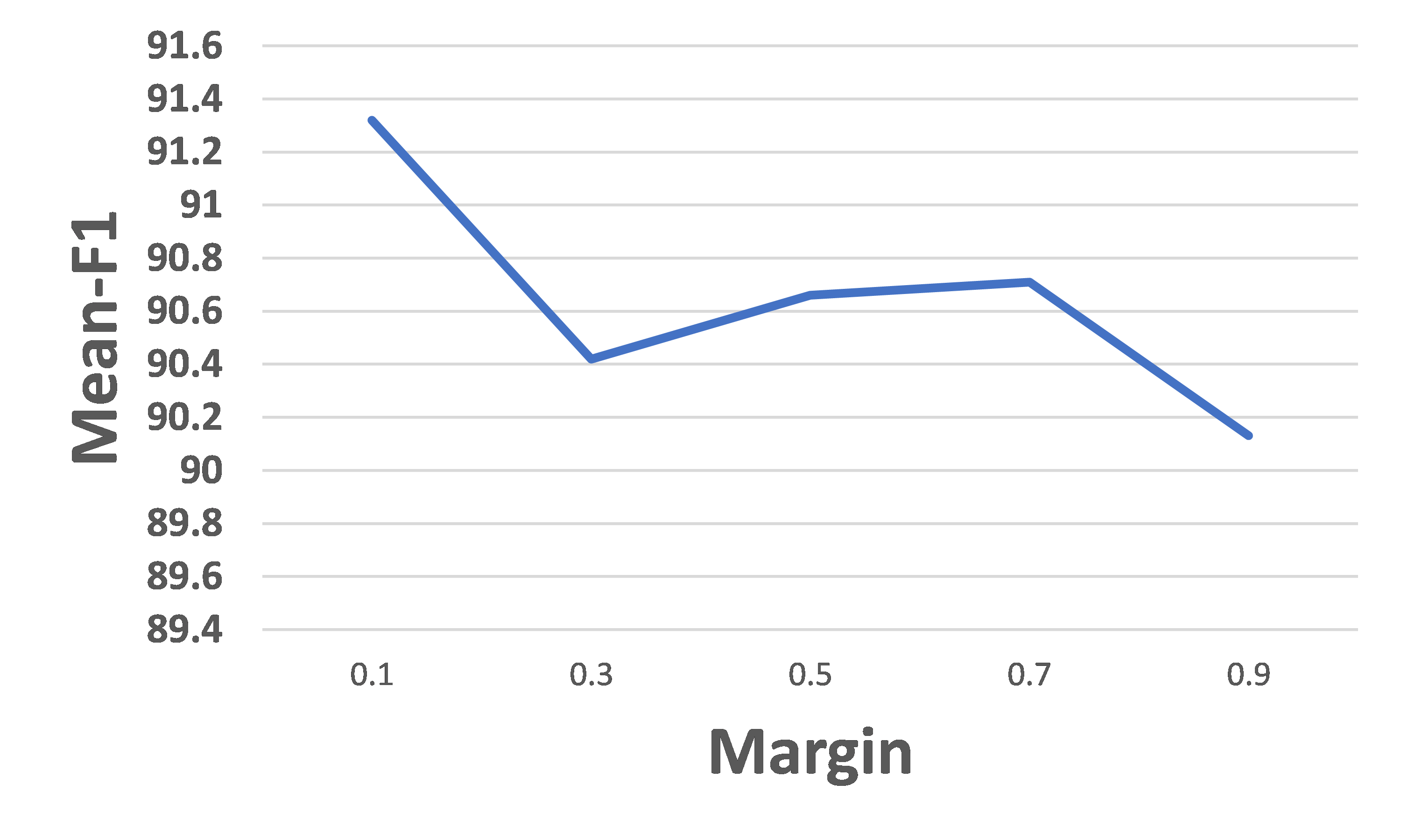}
        \caption{Robustness of the proposed PCCT with respect to the margin $\alpha$  on all three datasets (from left: Skin7, Skin198, ChestXray-COVID).}
        \label{fig:margin}
    \end{center}
\end{figure*}

Furthermore, we evaluated the proposed PCCT with different model architectures, varying sizes of output dimension, and varying values for the margin $\alpha$. From Table~\ref{tab:architectures}, it can be observed that with four different CNN models (ResNet-50~\cite{he2016deep}, DenseNet-121~\cite{huang2017densely}, Inception-v4~\cite{szegedy2017inception}, VGG-19~\cite{simonyan2014very}), the proposed PCCT consistently outperforms all the baseline methods on the dataset Skin198. Tests with varying output dimensions (Figure~\ref{fig:dimension}) also show that the proposed PCCT is still consistently better than corresponding baseline methods with the same model architecture. Furthermore, when varying the margin $\alpha$ in the 
class-center involved triplet loss,  the performance of the proposed PCCT is relatively stable on all three datasets (Figure~\ref{fig:margin}), e.g., 
the mean F1 score remained stably and varied only from $85.98\%$ to $86.39\%$ in the range $[0.1, 0.9]$ for $\alpha$ on Skin7. 
All the results are consistently supporting that the proposed two-stage PCCT method is stable and generalizable.


\subsection{Ablation study}

Here the effect of each stage in the proposed PCCT method was evaluated.
For the effect of the first stage, the tests on all three datasets (Table~\ref{tab:one_stage}) consistently support that the pre-trained feature extractor by the class-balanced triplet loss is crucial for the second-stage training by the class-center triplet loss. Compared to the only second-stage training (i.e., training the randomly initialized feature extractor by the class-center based triplet loss), the improvement from the two-stage training is particularly 
obvious on Skin198 and ChestXray-COVID datasets. For example, on the Skin198, with the help of the first stage training process, PCCT achieves the mean F1 score (MF1) of $65.20\%$, clearly higher than the one-stage method (MF1 58.89\%).
For the class-center involved loss in the second stage, the experiments show that the second stage further improves the performance of the class-balanced first-stage triplet loss (i.e., `Only first-stage' in Table~\ref{tab:one_stage}) to handle the class imbalance issue, where our PCCT achieves MF1 86.20\% on Skin7, clearly better than the result (MF1 84.02\%) from the original triplet loss. 

\begin{table*}[!ht]
	\begin{center}
	\caption{Effect of the two-stage training by the class-center based triplet loss. 
	} 
	\label{tab:one_stage}
{
	\begin{tabular}{c|ccc|ccc|ccc}
    \toprule
          \multirow{2}{*}{Methods}  & \multicolumn{3}{c}{Skin7} & \multicolumn{3}{c}{Skin198}& \multicolumn{3}{c}{ChestXray-COVID} \\
         \cmidrule(r){2-4}\cmidrule(r){5-7}\cmidrule(r){8-10}
         \multicolumn{1}{c|}{ }
          & MF1 & MCP &\multicolumn{1}{c}{MCR} & MF1 & MCP &\multicolumn{1}{c}{MCR} & MF1 & MCP &\multicolumn{1}{c}{MCR}\\
          \midrule
         
         Only first-stage & \tabincell{c}{84.02 \\ \scriptsize{(1.50)}} & \tabincell{c}{88.96  \\ \scriptsize{(1.85)}} & \tabincell{c}{80.44 \\ \scriptsize{(2.16)}} & 
         \tabincell{c}{64.25 \\ \scriptsize{(1.36)}} & \tabincell{c}{68.55 \\ \scriptsize{(1.17)} } & \tabincell{c}{64.39 \\ \scriptsize{(1.31)}} & 
         \tabincell{c}{81.59 \\ \scriptsize{(0.54)}} & \tabincell{c}{87.43 \\ \scriptsize{(2.40)}} & \tabincell{c}{77.37 \\ \scriptsize{(1.11)}}  \\
         
		Only second-stage & \tabincell{c}{85.98 \\ \scriptsize{(0.96)}} & \tabincell{c}{{87.89} \\ \scriptsize{(2.02)} } & \tabincell{c}{84.45  \\ \scriptsize{(0.40)}} & \tabincell{c}{58.89 \\ \scriptsize{(1.34)}} & \tabincell{c}{61.09 \\ \scriptsize{(1.54)}} & \tabincell{c}{61.11 \\ \scriptsize{(1.17)}} & \tabincell{c}{67.41 \\ \scriptsize{(4.97)}} & \tabincell{c}{68.80 \\ \scriptsize{(5.90)}} &\tabincell{c}{ 66.90\\ \scriptsize{(4.68)}} \\
			\midrule
			Two-stage (PCCT) & \tabincell{c}{{86.20} \\ \scriptsize{(1.07)}} & \tabincell{c}{87.77 \\ \scriptsize{(1.54)} } & \tabincell{c}{{84.98} \\ \scriptsize{(0.75)}} &
         \tabincell{c}{{65.20} \\ \scriptsize{(1.49)}} & \tabincell{c}{{68.40} \\ \scriptsize{(1.36)}} & \tabincell{c}{{66.02} \\ \scriptsize{(1.50)}} &
         \tabincell{c}{{90.66} \\ \scriptsize{(1.43)}} & \tabincell{c}{{92.97} \\ \scriptsize{(1.58)}} &\tabincell{c}{ {88.89}\\ \scriptsize{(2.70)}} \\

			\bottomrule
			\end{tabular}
		}
	\end{center}
\end{table*}




\subsection{Extensions to other metric learning}
The role of class centers in the triplet loss may also be extended to relevant metric learning, e.g., based on the pair-wise ranking loss and the quadruplet loss. As introduced in Section~\ref{sec:method_extend}, class centers can be easily used to replace the samples in the ranking loss and the quadruplet loss, resulting in the class-center based ranking loss and quadruplet loss respectively. 
As shown in Table~\ref{tab:double_quadruple}, compared to the original pair-wise ranking loss and the quadruplet loss, class-center involved losses can help train better feature extractors on all three datasets. This suggests that class centers may be potentially applied in various metric learning strategies where multiple samples as training units are involved. 

\begin{table*}[!ht]
    \centering
    \caption{Extension of the class-center based triplet loss to the class-center based pair-wise ranking loss and quadruplet loss. Class-center based loss performs better than the corresponding original loss on all three datasets.}
    \label{tab:double_quadruple}
    {
        \begin{tabular}{c|ccc|ccc|ccc}
    \toprule
     \multirow{2}{*}{Methods}  
          & \multicolumn{3}{c}{Skin7} & \multicolumn{3}{c}{Skin198}& \multicolumn{3}{c}{ChestXray-COVID} \\
         \cmidrule(r){2-4}\cmidrule(r){5-7}\cmidrule(r){8-10}
         \multicolumn{1}{c|}{ }
          & MF1 & MCP &\multicolumn{1}{c}{MCR} & MF1 & MCP &\multicolumn{1}{c}{MCR} & MF1 & MCP &\multicolumn{1}{c}{MCR} \\
         \midrule
         Pair-wise (original) & \tabincell{c}{{84.54} \\ \scriptsize{(0.71)}} & \tabincell{c}{{{88.95}} \\ \scriptsize{(1.15)}} & \tabincell{c}{{81.14} \\ \scriptsize{(1.18)}} & \tabincell{c}{{53.87} \\ \scriptsize{(1.61)}} &  \tabincell{c}{{60.37} \\ \scriptsize{(1.47)}} & \tabincell{c}{{52.59} \\ \scriptsize{(1.75)}}  & \tabincell{c}{{81.03} \\ \scriptsize{(1.14)}} & \tabincell{c}{{87.48} \\ \scriptsize{(3.45)}} & \tabincell{c}{{76.70} \\ \scriptsize{(3.31)}} \\
         Pair-wise (class centers) & \tabincell{c}{{{85.81}} \\ \scriptsize{(1.16)}} & \tabincell{c}{{88.70} \\ \scriptsize{(1.70)}} & \tabincell{c}{{{83.53}} \\ \scriptsize{(0.98)}} & \tabincell{c}{{{64.82}} \\ \scriptsize{(2.21)}} &  \tabincell{c}{{{67.51}} \\ \scriptsize{(2.33)}} & \tabincell{c}{{{66.06}} \\ \scriptsize{(2.13)}}  & \tabincell{c}{{{90.05}} \\ \scriptsize{(1.20)}} & \tabincell{c}{{{92.89}} \\ \scriptsize{(1.36)}} & \tabincell{c}{{{87.72}} \\ \scriptsize{(2.33)}} \\
         \midrule
         
         
         Quadruplet (original) & \tabincell{c}{{84.42} \\ \scriptsize{(1.32)}} &  \tabincell{c}{{{89.03}} \\ \scriptsize{(1.49)}} & \tabincell{c}{{81.15} \\ \scriptsize{(1.74)}} & \tabincell{c}{{63.08} \\ \scriptsize{(1.50)}} &  \tabincell{c}{{67.59} \\ \scriptsize{(1.95)}} & \tabincell{c}{{63.23} \\ \scriptsize{(1.58)}} & \tabincell{c}{{82.51} \\ \scriptsize{(1.71)}} &  \tabincell{c}{{90.36} \\ \scriptsize{(2.10)}} & \tabincell{c}{{77.19} \\ \scriptsize{(2.63)}}\\
         Quadruplet (class centers) & \tabincell{c}{{{86.11}} \\ \scriptsize{(1.21)}} &  \tabincell{c}{{88.79} \\ \scriptsize{(1.53)}} & \tabincell{c}{{{84.08}} \\ \scriptsize{(1.35)}} & \tabincell{c}{{{65.00}} \\ \scriptsize{(1.83)}} &  \tabincell{c}{{{67.77}} \\ \scriptsize{(1.72)}} & \tabincell{c}{{{66.11}} \\ \scriptsize{(1.99)}} & \tabincell{c}{{{90.36}} \\ \scriptsize{(1.05)}} &  \tabincell{c}{{{{92.02}}} \\ \scriptsize{(1.83)}} & \tabincell{c}{{{88.99}} \\ \scriptsize{(2.29)}}\\
         
         \bottomrule
    \end{tabular}
    }
\end{table*}




\subsection{Discussions}
From the above extensive evaluations, it is clear that the proposed two-stage learning framework with certain metric learning loss is effective in handling the class-imbalance issue and outperforms widely used strong baselines under various conditions. This is consistent with the previously reported two-stage learning strategy TSD for the class imbalance issue~\cite{Kang2020DecouplingRA}. However, different from TSD which simply trains the model (mainly the feature extractor) with cross-entropy loss at the first stage and then applies certain class re-balancing strategy at the second stage, the class-balanced triplet loss and the class-center involved triplet loss in the proposed framework can further help train a better feature extractor by enforcing more compact within-class distribution and enlarging the separation between classes. 
Note that the class centers in the proposed framework can be learned together with model parameters, reducing the computational cost for class center estimate during model training. Consistent with our study, one recent work~\cite{Cui_2021_ICCV} 
used class center loss to help train a feature extractor for imbalanced image classification tasks.
In addition, other types of metric learning losses could be applied as well in the two-stage learning framework, e.g., using contrastive loss~\cite{chopra2005learning} in the first stage, which will be part of our future exploration following this study.

The proposed framework focuses on training a class-balanced feature extractor. Therefore, it is complementary to many existing strategies which focus on the input side (e.g., re-sampling) or output side (re-weighting, focal loss, etc.) of the classifier. For example, our method may be combined with existing strategies to further enlarge separation between classes, e.g., with the help of the distribution-aware margin loss~\cite{Cao2019LDAM}, 
or combined with data augmentation techniques like Mixup to further alleviate the data imbalance between classes~\cite{zhong2021improving}. 

Although promising performance has been achieved on various imbalanced medical datasets, the proposed two-stage framework with the triplet loss has an obvious limitation, i.e., relatively longer training time compared to the single-stage methods. 
Although the proposed Efficient-PCCT can decrease training time in the first stage, the second stage is still relatively time consuming. Replacing the triplet loss by other types of metric learning loss (e.g., contrastive loss) could largely alleviate this issue. However, it is worth noting that the inference time is determined mainly based on the model backbone and therefore inference is in general near real-time for medical diagnosis.


\section{Conclusion}
In this paper, we propose a two-stage method PCCT to handle the class imbalance issue. 
PCCT consists of two novel training stages.
The triplet loss with the class-balanced triplet sampler is proposed to optimize the feature extractor model in the first stage, and then the class-center involved triplet loss is proposed to further fine-tune the feature extractor in the second stage such that the distribution of each class in the feature space becomes more compact and easily separated from each other.
PCCT outperforms the widely used strong baselines and achieves state-of-the-art performance on three medical image classification tasks.
The stability and generality of the proposed PCCT on various model backbones, output sizes, and hyperparameter settings are demonstrated by extensive experiments.
The class-center idea has also been easily extended to other relevant metric learning approaches. We expect that this two-stage method will help effectively develop intelligent diagnosis systems for both common and rare diseases. 




\bibliographystyle{IEEEtran}
\bibliography{pcct.bib}

\end{document}